\documentclass[10pt,twocolumn,letterpaper]{article}

\usepackage{cvpr}              

\usepackage[dvipsnames]{xcolor}

\definecolor{cvprblue}{rgb}{0.21,0.49,0.74}

\usepackage[pagebackref,breaklinks,colorlinks,citecolor=cvprblue]{hyperref}
\usepackage{multirow}
\usepackage{bm}
\usepackage{xspace}
\usepackage{comment}
\usepackage{booktabs}

\definecolor{asparagus}{rgb}{0.53, 0.66, 0.42}

\newcommand{\jedi}{\texttt{JeDi}\xspace}
\usepackage{pifont}
\newcommand{\cmark}{\color{asparagus}\ding{51}}%
\newcommand{\xmark}{\color{red}\ding{55}}%
\usepackage[linesnumbered,ruled]{algorithm2e}
\usepackage{float}
\usepackage{algorithmic}

\def\paperID{2603} 
\def\confName{CVPR}
\def\confYear{2024}

\title{JeDi: Joint-Image Diffusion Models for Finetuning-Free Personalized Text-to-Image Generation}

\author{%
  Yu Zeng$^{1,3}$\quad Vishal M.~Patel$^{1}$ \quad Haochen Wang
$^{2}$\quad Xun Huang$^{3}$\quad \\ Ting-Chun Wang$^{3}$\quad Ming-Yu Liu$^{3}$\quad Yogesh Balaji$^{3}$\quad \\
  {\small$^{1}$Johns Hopkins University\quad $^{2}$TTI-Chicago  \quad $^{3}$NVIDIA Research}
}

\begin{document}
\twocolumn[{
\renewcommand\twocolumn[1][]{#1}%
\maketitle
\begin{center}
    \centering
    \vspace{-1pt}
    \includegraphics[width=\textwidth]{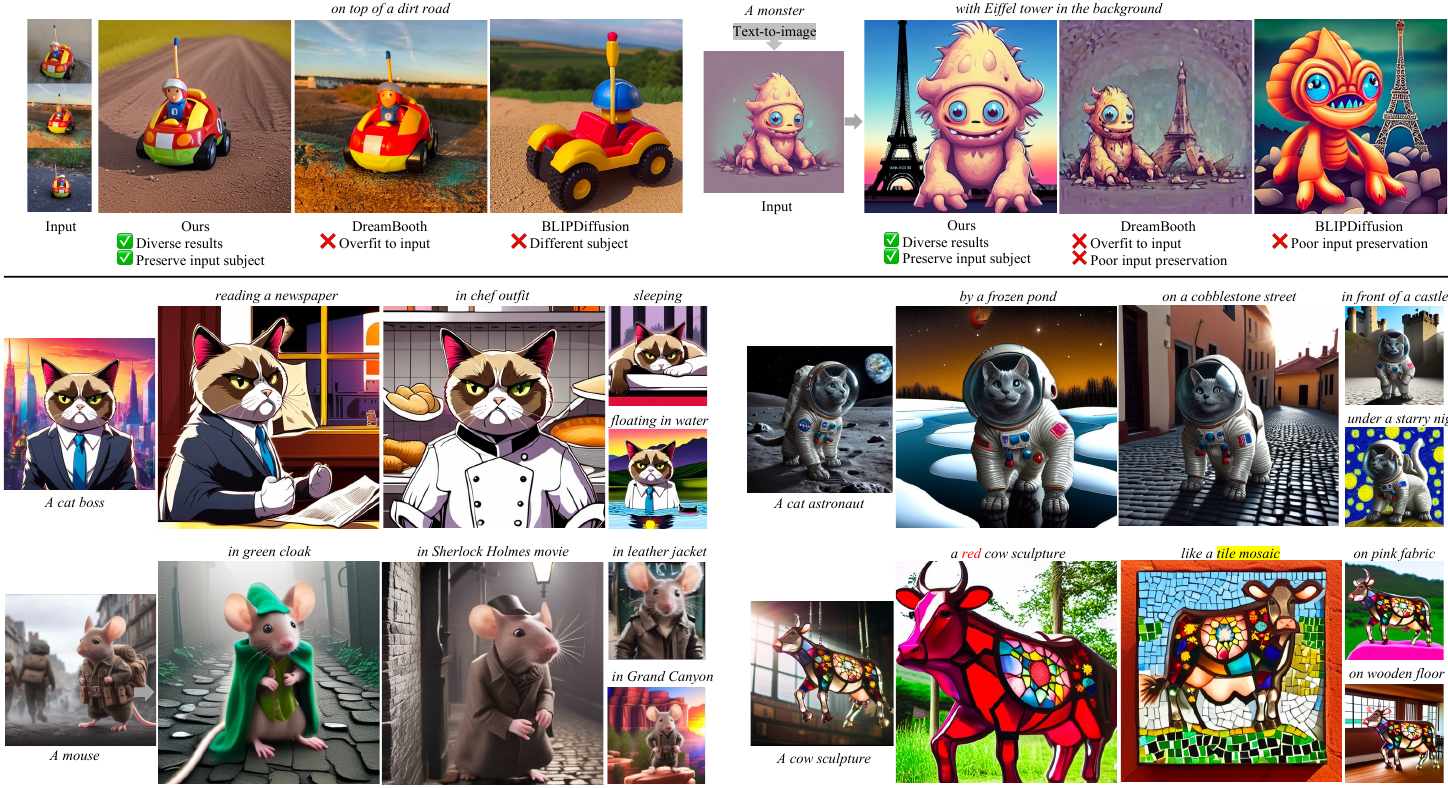}\\
    \captionof{figure}
    {We present Joint-Image Diffusion (\jedi), a finetuning-free image personalization model that can operate on any number of reference images. \jedi is able to preserve the appearance of custom subjects while generating novel variations. As shown in the top row, \jedi does not suffer from the issues of overfitting and lack of diversity exhibited by the prior models. The examples in the bottom two rows demonstrate \jedi's high-quality results on challenging personalization tasks.}
    \label{fig0}
\end{center}
}]

\begin{abstract}
Personalized text-to-image generation models enable users to create images that depict their individual possessions in diverse scenes, finding applications in various domains. To achieve the personalization capability, existing methods rely on finetuning a text-to-image foundation model on a user's custom dataset, which can be non-trivial for general users, resource-intensive, and time-consuming. Despite attempts to develop finetuning-free methods, their generation quality is much lower compared to their finetuning counterparts. In this paper, we propose Joint-Image Diffusion (\jedi), an effective technique for learning a finetuning-free personalization model. Our key idea is to learn the joint distribution of multiple related text-image pairs that share a common subject. To facilitate learning, we propose a scalable synthetic dataset generation technique. Once trained, our model enables fast and easy personalization at test time by simply using reference images as input during the sampling process. Our approach does not require any expensive optimization process or additional modules and can faithfully preserve the identity represented by any number of reference images. Experimental results show that our model achieves state-of-the-art generation quality, both quantitatively and qualitatively, significantly outperforming both the prior finetuning-based and finetuning-free personalization baselines. Project page https://research.nvidia.com/labs/dir/jedi
\vspace{-5pt}
\end{abstract}    
\section{Introduction}
\label{sec:intro}

The state-of-the-art in text-to-image generation has advanced significantly in the last two years, propelled by the emergence of large-scale diffusion models and paired image-text datasets~\cite{ rombach2022high,balaji2022ediffi,saharia2022photorealistic,ramesh2022hierarchical, podell2023sdxl, dai2023emu, Betker2023ImprovingIG}. Despite their superior capability in generating high-quality images well-aligned to the input text prompts, existing models cannot generate novel images depicting specific custom objects or styles that are only available as few reference images external to the training datasets. To address this important use case, various personalization methods have been developed.

The key challenge of personalized image generation is to produce distinct variations of a custom subject while preserving its visual appearance. Most existing approaches achieve this goal by finetuning a pre-trained model on a reference set of images to make it memorize the custom concept. Although these methods can yield good synthesis results, they require substantial resources and a long training time to fit the custom subject, and more than one reference image is needed to avoid overfitting. To overcome these challenges, there has been recent interest in developing \emph{finetuning-free personalization} methods~\cite{xiao2023fastcomposer,wei2023elite,li2023blip,shi2023instantbooth}. These methods typically encode the reference images into a compact feature space, and condition the diffusion model on the encoded features. However, the encoding step results in information loss, leading to poor appearance preservation, especially for challenging unusual objects as seen in Fig.~\ref{fig0}. Therefore, the performance of encoder-based personalization techniques is inferior to finetuning-based approaches. 

\begin{figure*}[h]
    \begin{center}
    \vspace{-10pt}
    \includegraphics[width=\linewidth]{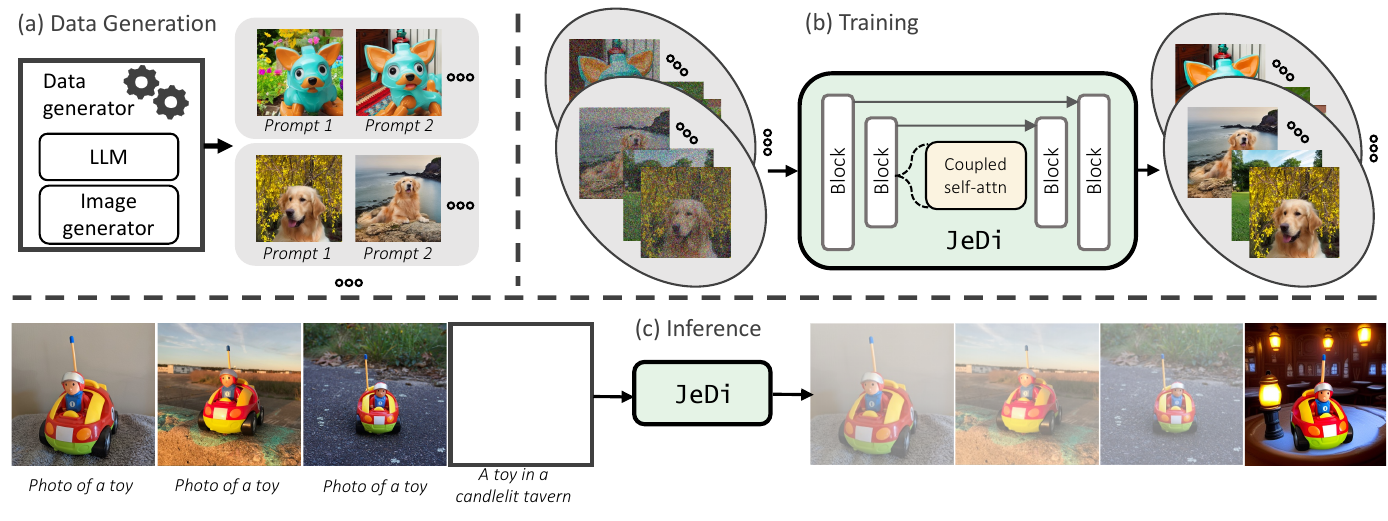}
    \end{center}
    \vspace{-20pt}
    \caption{ \textbf{Overall framework.} (a) We generate training data by using large language models and prompting pretrained single-image diffusion models. (b) During training, the \jedi model learns to denoise multiple same-subject images together, where each image attends to every image of the same subject set through coupled self-attention. (c) At inference, personalized generation is performed in an inpainting fashion where the goal is to generate the missing images of the joint-image set. }
    \vspace{-10pt}
    \label{fig_motiv}
\end{figure*}

In this paper, we present \jedi, a novel approach for finetuning-free personalized text-to-image generation that excels at preserving input reference content. Our core idea is to train a diffusion model to learn a joint distribution of multiple related text-image pairs that share a common subject. As illustrated in Fig.~\ref{fig_motiv}, this goal is achieved using two key ingredients: First, we construct a synthetic dataset of related images in which each sample contains a set of text-image pairs that share a common subject. We present a scalable approach for creating such a dataset using LLMs and pre-trained single-image diffusion models. Second, we modify the architecture of existing text-to-image diffusion models to encode relationships between multiple images in a sample set. Specifically, we adapt the self-attention layers of the diffusion U-Net so that the attention blocks corresponding to different input images are coupled. That is, the self-attention layer corresponding to each image co-attends to every other image in the sample set. The use of the coupled self-attentions at different levels of hierarchy in the U-Net provides a much stronger representation needed for good input preservation.

At test time, \jedi can take multiple text prompts as input and generate images of the same subject in different contexts. By simply substituting reference images as observed variables in the sampling process, \jedi can generate personalized images based on any number of reference images. We utilize guidance techniques~\cite{ho2021classifier} on reference images to further improve the image alignment. \jedi can achieve high-fidelity personalization results even in challenging cases involving unique subjects (Fig.~\ref{fig0}, \ref{fig:our_generations_highres}), using as few as a single reference image.

Our \textbf{key contributions} are summarized as follows: 
\begin{itemize}[noitemsep]
\item We propose a finetuning-free text-to-image generation method with a novel joint-image diffusion model.
\item We present a simple and scalable data synthesis pipeline for generating a multi-image personalization dataset with images sharing the same subject. 
\item We design novel architecture and sampling techniques such as coupled self-attention and image guidance for achieving high-fidelity personalization.

\end{itemize}


\section{Related Work}
\label{sec:related}
\subsection{Text-to-Image Generation}


Denoising diffusion models~\cite{dhariwal2021diffusion,ho2020denoising,song2020score} formulate the image generation task as a series of progressive denoising steps. The denoising network can be trained conditioned on text embeddings to generate images from an input caption. 
 DALL-E2~\cite{ramesh2022hierarchical} achieves high-resolution text-to-image synthesis using two diffusion models: the first model transforms a CLIP text embedding to a CLIP image embedding, while the second model transforms the image embedding to an output image. Imagen~\cite{saharia2022photorealistic} trains a cascaded diffusion model conditioned on T5 language embeddings~\cite{raffel2020exploring}. eDiff-I~\cite{balaji2022ediffi} uses an ensemble of expert denoisers to increase the model capacity, with each expert specializing in a specific noise range. Latent diffusion models~\cite{rombach2022high, podell2023sdxl, dai2023emu}  train the diffusion model in a compact latent space of an autoencoder for efficient training and sampling.

\subsection{Personalized Text-to-Image Generation}
\begin{table}[t]
\begin{center}
\caption{In contrast to prior work, \jedi does not require finetuning or the use of image encoders for image personalization.  }
\vspace{-5pt}
\resizebox{\linewidth}{!}{%
\begin{tabular}{ l c c}
\toprule
  Method & Finetuning-free & Encoder-free \\
  \hline 
  DreamBooth~\cite{ruiz2023dreambooth} & \xmark & \cmark \\
CustomDiffusion~\cite{kumari2022customdiffusion} & \xmark & \cmark\\
  ELITE~\cite{wei2023elite} & \cmark & \xmark \\
    BLIPDiffusion~\cite{li2023blip} & Optional & \xmark \\
      \jedi & \cmark & \cmark \\
\bottomrule
\end{tabular}
}
\end{center}
\vspace{-25pt}
\end{table}
\textbf{Finetuning based methods. }Most prior works achieve image personalization by finetuning the diffusion model on a custom dataset. Dreambooth~\cite{ruiz2023dreambooth} finetunes the entire model weights on the reference set, with a loss on images of similar concepts as regularization. CustomDiffusion~\cite{kumari2022customdiffusion} optimizes only a few parameters to enable fast tuning, and combines multiple finetuned models for multi-concept personalization. Textual Inversion~\cite{gal2022image} projects the reference images onto the text embedding space through an optimization process.  
SVDiff~\cite{han2023svdiff} finetunes only the singular values of the weight matrices to reduce the risk of overfitting. These finetuning-based methods require a substantial amount of resources and long training time, and often need multiple reference images per custom subject.

\textbf{Finetuning-free methods. }To improve the efficiency of image customization, there has been recent interest in developing finetuning-free methods. These approaches typically use an image encoder to encode a reference image onto a compact feature space, and train the diffusion model conditioned on this feature vector. BLIPDiffusion~\cite{li2023blip} uses BLIP-2~\cite{li2023blip2} encoder, while FastComposer~\cite{xiao2023fastcomposer} uses a CLIP~\cite{radford2021learning} encoder for image encoding. ELITE~\cite{wei2023elite} and InstantBooth~\cite{shi2023instantbooth} use a learnable image encoder trained jointly with the diffusion model. These encoder-based methods produce reasonable results for common subjects, but often fail to generate uncommon subjects and preserve fine-grained details due to the information loss in the encoding step. In contrast, our approach directly trains a joint-image diffusion model without an encoding step, resulting in better input preservation even for challenging objects.


\textbf{Personalization dataset. }To achieve finetuning-free personalization, a training dataset comprised of same-subject image sets is required. Methods like~\cite{li2023blip,xiao2023fastcomposer,shi2023instantbooth,wei2023elite} rely on image augmentation and background removal to construct training data, which often does not provide sufficient variations for the same subject. To improve the diversity, we present a scalable data generation approach by prompting pre-trained single-image diffusion models to produce multi-image photo collages with good variation. 
\label{sec:method}
\section{Method}

\begin{figure*}[t]
    \begin{center}
    \vspace{-10pt}
    \includegraphics[width=\linewidth]{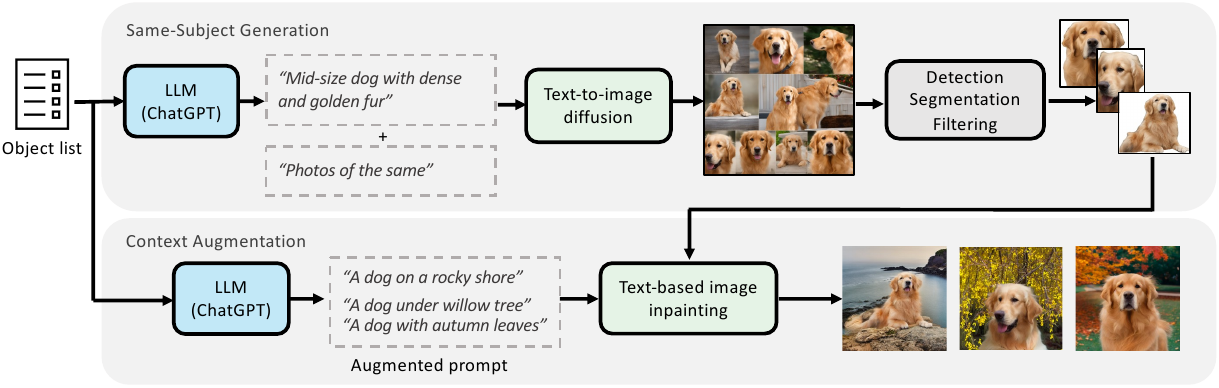}
    \end{center}
    \vspace{-15pt}
    \caption{\textbf{Data generation process.} We construct Synthetic Same-Subject (S$^3$) dataset by first prompting the pretrained text-to-image diffusion models to generate same-subject photo collages, and then increasing the diversity using text-based background inpainting.}
    \label{fig_data}
\end{figure*}
\begin{figure}[t]
    \begin{center}
    \vspace{-10pt}
    \includegraphics[width=\linewidth]{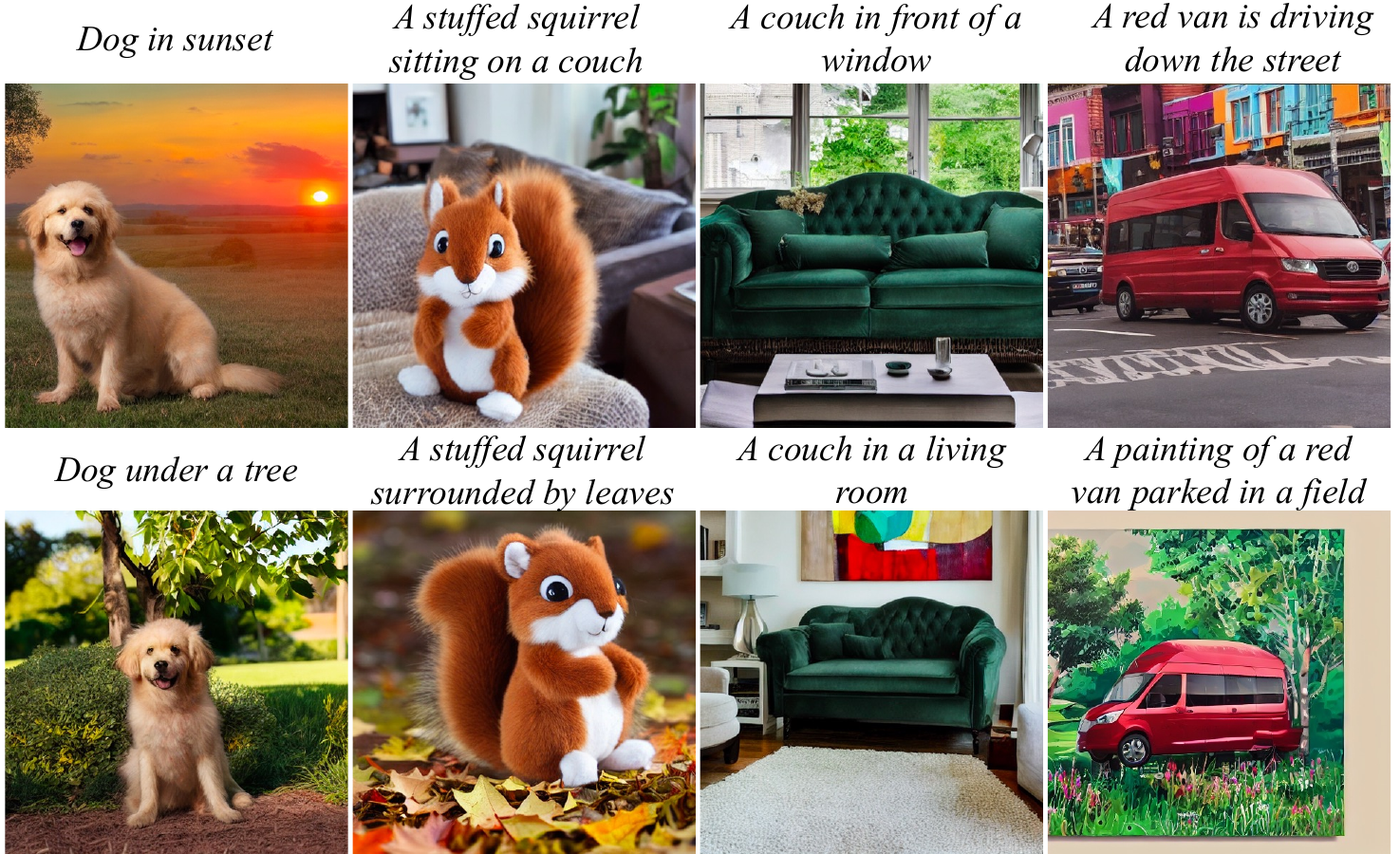}
    \end{center}
    \vspace{-15pt}
    \caption{\textbf{Samples from the synthetic same-subject (S$^3$) dataset.} Each column denotes different images from one joint-data sample.}
    \label{fig_data_images}
    \vspace{-10pt}
\end{figure}


\subsection{Dataset Creation}
\label{sec:data}
Training a model to produce a joint distribution of multiple same-subject images requires a dataset where each sample is a set of images sharing a common subject. While there exist some same-subject datasets such as CustomConcept101~\cite{kumari2022customdiffusion} and DreamBooth~\cite{ruiz2023dreambooth}, they are small in scale and lack sufficient variations desired for diffusion model training. Therefore, we create a diverse large-scale dataset of image-text pairs containing the same subject, called the Synthetic Same-Subject (S$^3$) dataset, using large language models~\cite{chatgpt} and single-image diffusion models~\cite{podell2023sdxl}.

 Fig.~\ref{fig_data} illustrates our data generation process. We first start with a list of common objects and prompt ChatGPT to generate a text description for each object in the list. Then, we use the pre-trained SDXL~\cite{podell2023sdxl} model to generate a dataset of same-subject photo collages by appending the text \emph{``photos of the same"} to each of the text prompt generated in the previous step. We observed that by prompting the SDXL model this way, it can generate photo collages of the same subject with varying poses. However, the generated images usually contain a close-up view of an object in a simple background. To increase the data diversity, we employ a post-processing step that performs background augmentation on the generated objects. 
 
 Given a generated photo collage, we first run object detection~\cite{liu2023grounding} and segmentation~\cite{kirillov2023segment} to separate out object instances and extract foreground region. We discard pairs of instances with CLIP~\cite{radford2021learning} image scores lower than $0.95$ as they may not contain the same subject. We then paste the object at a random location in a blank image, and use the stable diffusion inpainting model to inpaint the background based on a new prompt related to the same object name. In addition, we use InstructPix2Pix~\cite{brooks2023instructpix2pix} to stylize the generated samples with a probability of $0.5$ to increase style variation using randomly selected style prompts. Fig.~\ref{fig_data_images} shows some examples of the text-image data generated through this process. The generated samples have consistent subjects with good diversity and pose variations. 

\subsection{Joint-Image Diffusion}
\label{sec:model_joint}

\begin{figure*}[t]
    \begin{center}
    \vspace{-10pt}
    \includegraphics[width=\linewidth]{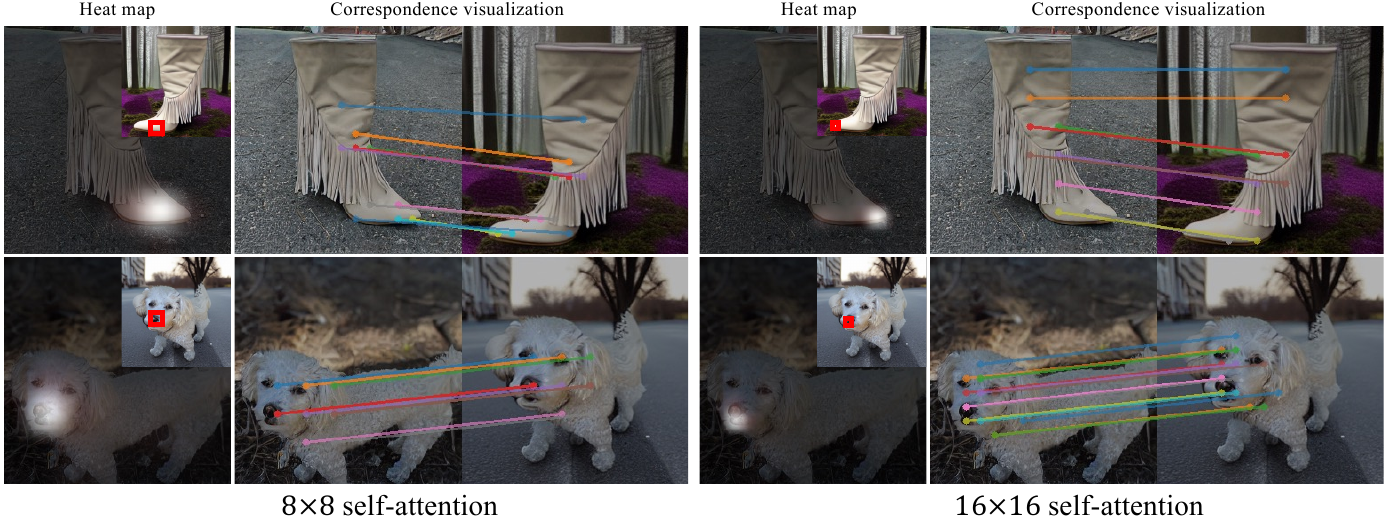}\\
    \end{center}
    \vspace{-15pt}
    \caption{\textbf{Visualization of the coupled self-attentions.} 
    For both scales (8x8 and 16x16), the correspondence map (Corr.) shows the connections with the highest weights between elements in the two images. 
    The heatmap visualizes the distribution of the attention weights in an image for a specific element in another image (marked with a red box). We observe that similar regions in different images are co-attended in the coupled self-attention layers.  
    }
    \vspace{-10pt}
    \label{fig_attn_vis}
\end{figure*}

\begin{figure}[t]
    \begin{center}
    \includegraphics[width=\linewidth]{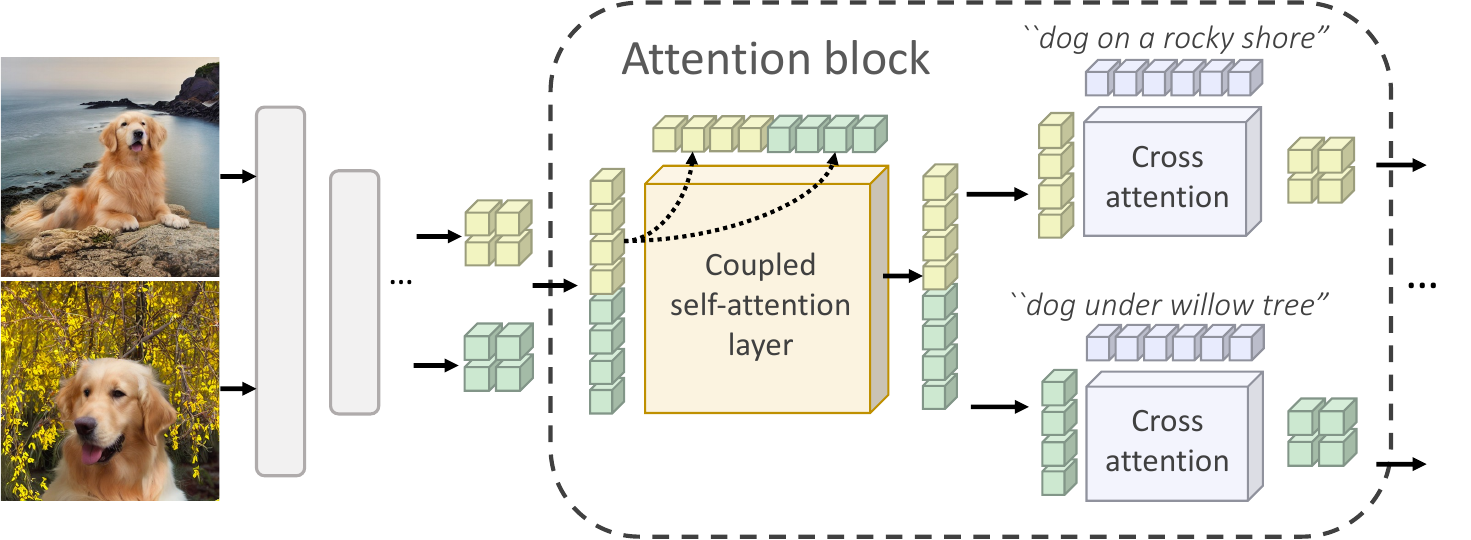}
    \end{center}
    \vspace{-10pt}
    \caption{\textbf{Attention architecture. }In a coupled self-attention block, features corresponding to each spatial location attends to every location across all images in the image set. Following the coupled self-attention block, features of each individual image attends to their respective text embedding in cross-attention layers. }
    \vspace{-10pt}
    \label{fig_model}
\end{figure}

The goal of training a joint-image diffusion model is to generate multiple related images sharing the same subject. Conventional diffusion models~\cite{rombach2022high,saharia2022photorealistic,nichol2021glide,balaji2022ediffi}, however, can only generate individual images independently as the network architecture does not have any connections between different samples in a batch. 
We found that with a simple modification, a single-image diffusion model can be adapted to a joint-image model which can generate images having related content (such as the same custom subject). 

More specifially, given a set of same-subject noisy input images, we modify the attention layers of the U-Net to fuse together the self-attention features for different images in the same set. As illustrated in Fig.~\ref{fig_model}, a coupled self-attention layer has features at each spatial location attending to every other location across all images in the set. 
Since the U-Net architecture has attention layers at various resolutions, the use of coupled self-attentions at multiple resolutions makes the generated image set to have consistent high-level semantic features as well as low-level attributes. Fig.~\ref{fig_attn_vis} visualizes the pixel-wise correspondences and attention heat maps of coupled self-attention layers. We observe that the co-attended regions across different images form the right correspondences across all resolutions. 


After a coupled self-attention layer, the output is fed to a regular cross-attention layer, which aligns the visual feature of each image to the corresponding text prompt. The coupled self-attention layer can be implemented by simply adding two reshaping operations before and after a regular self-attention, thus enabling simple and easy adaptation from pre-trained single-image diffusion models.


Fig.~\ref{fig_motiv}~(b) illustrates the training process. We start by creating noisy same-subject data by adding isotropic Gaussian noise, and train the joint-image diffusion model to denoise the data. Ideally, there is no limit on the size of each image set of the same subject. In our experiments, we randomly set the size to $2$, $3$, or $4$ during training. The training loss of \jedi is very similar to that of a regular diffusion model. We use $\bm{\epsilon}$-prediction and a simplified training objective introduced in~\cite{ho2020denoising}. The loss function is as follows,
\begin{equation}
L = \mathbb{E}_{\bm{\epsilon}\sim \mathcal{N}(0,1),t\sim[1,T]} \left [ \| \bm{\epsilon} - \bm{\epsilon_\theta} (\bm{x}_t)\|_2^2 \right ],
\end{equation}
where $\bm{\epsilon_\theta}$ represents the network parameterized by $\bm{\theta}$, $T$ is the number of diffusion steps, $\bm{x}_t$ is the $t$-step noised image set of size $N$,~\ie $\bm{x}=[x^1,x^2,...,x^N]$. We omit the text and timestep conditioning in $\bm{\epsilon_\theta}(\cdot)$ for simplicity. 
After training, the \jedi model can take multiple text prompts as input and generate images containing the same subject. 

\subsection{Personalized Text-to-Image Generation}\label{sec_personalize}
\noindent\textbf{Personalization as inpainting.} While the joint-image diffusion model discussed in the previous section can generate same-subject images, it does not input a reference image that needs to be personalized. In this section, we propose to solve the input image personalization problem by casting it as an inpainting task. That is, given a few text-image pairs as reference, the task of generating a new personalized sample can be viewed as inpainting the missing images of a joint-image set containing reference images (Fig.~\ref{fig_motiv}~(c)).


\setlength{\tabcolsep}{0.5pt}
\renewcommand{\arraystretch}{0.5}
\begin{figure*}[ht!]
    \centering
    \vspace{-10pt}
    \includegraphics[width=\textwidth]{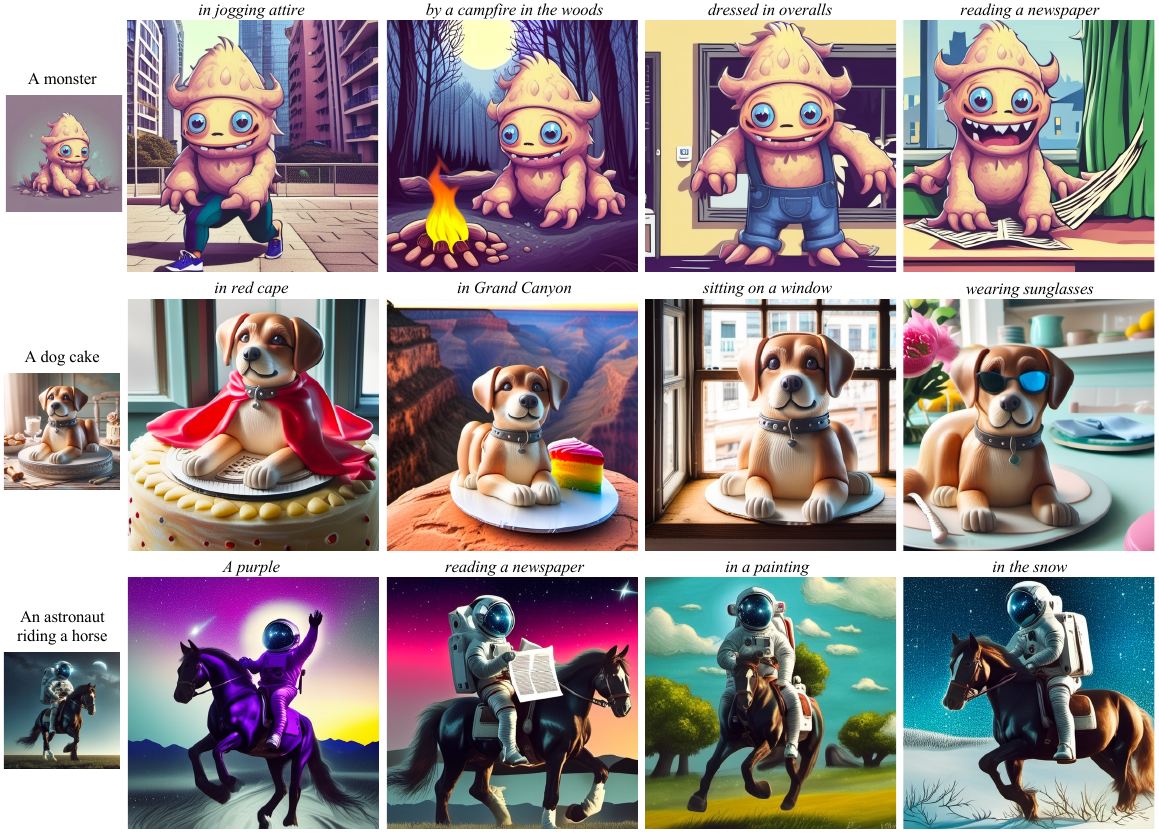}



    \vspace{-5pt}
    \caption{\textbf{Personalized text-to-image generations.} We show the generations obtained by our \jedi model on challenging uncommon input subjects shown in the left column. \jedi accurately preserves the reference image content while being faithful to the text prompt.}
    \vspace{-10pt}
    \label{fig:our_generations_highres}
\end{figure*}


We design the inpainting model by modifying the input layer of the diffusion U-Net so that it can be conditioned on reference images. More specifically, the input to the diffusion model is a concatenated list of noisy images, reference images and a binary mask indicating whether the reference image is used or not. When the binary mask is all $0$'s, the reference image is used. On the other hand, when the binary mask is all $1$'s, the reference image is an empty black image (indicating the missing images that need to be generated). During training, for every image in the joint-image set, we use the reference image with a probability of $0.5$ i.e., we assign the binary mask to $0$ with probability $0.5$. 
The training training loss can then be written as follows,
\begin{equation}
L = \mathbb{E}_{\bm{\epsilon}\sim \mathcal{N}(0,1),t\sim[1,T],\bm{m}\sim \binom {N} {0.5}} \left [ \| \bm{\epsilon} - \bm{\epsilon_\theta} (\bm{x}_t,\bm{\hat{x}}, \bm{M})\|_2^2 \right ],
\end{equation}
where $\bm{M}$ is the spatially repeated tensor of a binomial variable $\bm{m}$; $\bm{\hat{x}}=\bm{x}\odot\bm{M}$ denotes the reference images, with the unknown elements set to zero. 

During inference, we utilize the replacement trick~\cite{lugmayr2022repaint,song2020score} in which the known part of the joint-image set is replaced with the forward diffusion response of the clean reference image. 
Let $\hat{\bm{x}}=[\hat{x}^1, \hat{x}^2, ..., \hat{x}^n]$ be the reference images, which are the known elements in the image set $\bm{x}=[\hat{x}^1,...,\hat{x}^n,x^{n+1},...,x^N]$ to be generated.
During sampling, at each diffusion step $t$, we only keep the backward diffusion output for the unknown elements $x^{n+1},...,x^N$ while replacing the known part with the forward diffusion output,~\ie the noised real images $\hat{x}_t^1,...,\hat{x}_t^n$. 

\textbf{Image guidance. }
Classifier-free guidance is a popular technique used in single-image diffusion models to make the image generations more aligned to the input conditioning~\cite{ho2021classifier}. To improve the faithfulness of the generated samples to the input reference images, we use image guidance in addition to the text guidance during sampling. The score function with the use of image guidance is as follows:
\label{eq:guidance}
\begin{equation}
\tilde{\bm{\epsilon}}(\bm{x}_t,\bm{\hat{x}},\bm{M})=\bm{\epsilon}^0 + \lambda[\bm{\epsilon}_\theta(\bm{x}_t,\hat{\bm{x}},\bm{M})-\bm{\epsilon}^0]
\end{equation}
where $\bm{\epsilon}^0=\bm{\epsilon_\theta}(\bm{x}_t,\bm{0},\bm{M})$ represents the unconditional score when the text prompt and all reference images are set to null; $\lambda$ is the guidance scale. We find that the use of image guidance can significantly improve the fidelity to the input reference images. 
\section{Experiments}
\begin{figure*}[t]
    \begin{center}
    \includegraphics[width=\linewidth]{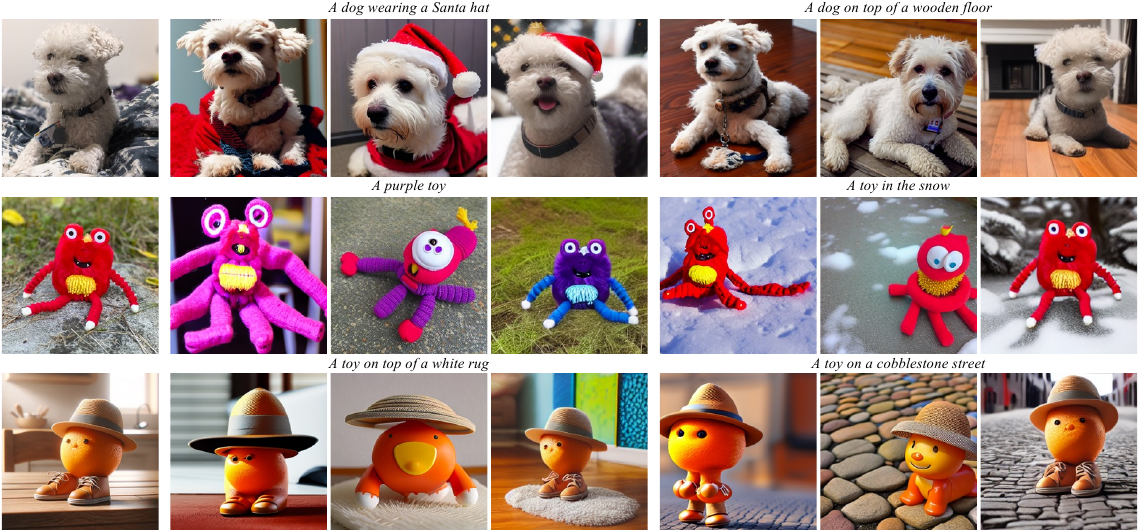}\\
    \scriptsize{ \hfill {Input} \hfill\hfill {ELITE} \hfill\hfill {BLIPD} \hfill\hfill {Ours\textcolor{white}{1}} \hfill\hfill {ELITE} \hfill\hfill {BLIPD} \hfill\hfill {Ours\textcolor{white}{1}} \hfill\hfill}
    \end{center}
    \vspace{-15pt}
    \caption{\textbf{Visual comparison with finetuning-free methods.} \jedi can faithfully preserve the details of input content even for challenging uncommon objects (row 2 and 3). ELITE and BLIPD fail in such cases and can get good results only in common object classes (row 1).}
    \vspace{-0pt}
    \label{fig_cmp}
\end{figure*}
\begin{figure*}[t]
    \begin{center}
    \includegraphics[width=\linewidth]{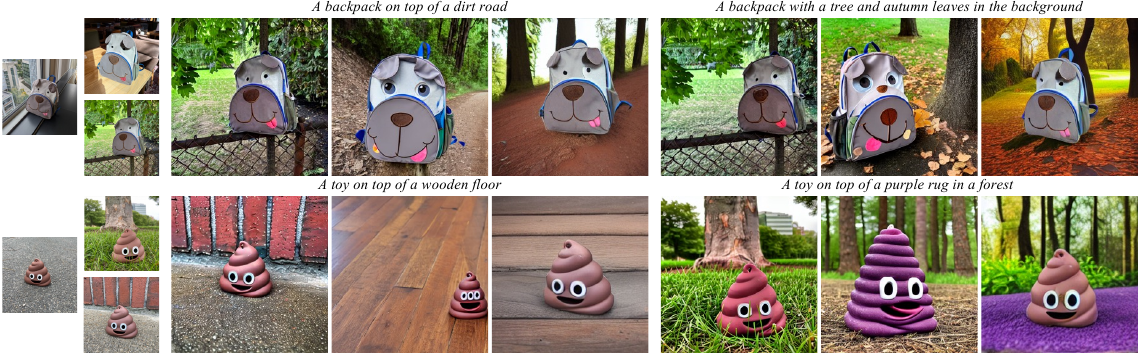}\\
    \scriptsize{ \hfill {Input} \hfill\hfill {DB} \hfill\hfill {CD} \hfill\hfill {Ours} \hfill\hfill {DB} \hfill\hfill {CD} \hfill\hfill {Ours} \hfill\hfill}
    \end{center}
    \vspace{-15pt}
    \caption{\textbf{Visual comparison with finetuning methods.} Finetuning methods memorize the input images (DB outputs) or result in poor input preservation (CD outputs) when the number of reference images is small. On the other hand, \jedi can generate images that are faithful both to the input images and the text prompt. }
    \vspace{-10pt}
    \label{fig_cmp2}
\end{figure*}

\textbf{Dataset. }We construct the Synthetic Same-Subject dataset (S$^3$ dataset) for training as described in Sec.~\ref{sec:data}. After CLIP filtering, we obtain $1.6M$ sets of images with each set containing $2$-$4$ images. We also include the video frames from WebVid10M~\cite{Bain21} and rendered multi-view images from Objaverse~\cite{deitke2023objaverse} during training, as the frames from the video and the rendered images from the asset usually have the same subject. We use the original video caption or asset caption as text prompts for all images obtained from the same video/asset. Additionally, we include the single-image data from LAION aesthetic dataset~\cite{schuhmann2021laion} and use a set size of $1$ for these images. To evaluate our models, we use the test dataset proposed in DreamBooth~\cite{ruiz2023dreambooth}. DreamBooth test set contains $30$ real-world subjects with $4$-$6$ images and $25$ prompts for each subject. 

\begin{table*}
\renewcommand*{\arraystretch}{1}
\vspace{-10pt}
\caption{\textbf{Quantitative comparisons.} All finetuning-based methods use 3 input images, while the finetuning-free methods use 1 input image. \jedi model with 1 input image outperforms all finetuning-free baselines, while the \jedi model with 3 input images outperforms all finetuning baselines. \jedi obtains a much higher masked DINO scores, which suggests that we achieve stronger input subject preservation compared to other baselines.
}
\vspace{-15pt}
\label{tab:comp}
\begin{center}
\setlength{\tabcolsep}{5pt}
\resizebox{0.8\linewidth}{!}{%
\begin{tabular}{c|l|cccccc}
\toprule
 & Method &CLIP-T ($\uparrow$) & CLIP-I ($\uparrow$) &MCLIP-I ($\uparrow$) & DINO ($\uparrow$) & MDINO ($\uparrow$)\\
\midrule
\multirow{ 2}{*}{Finetuning-based}  & DreamBooth~\cite{ruiz2023dreambooth} & 0.2812 & 0.8135&0.8683&0.6341&0.7115\\
& Custom Diffusion~\cite{kumari2022customdiffusion} & 0.3015 & 0.7952&0.8640&0.6343&0.7109
\\
\midrule
\multirow{ 4}{*}{Finetuning-free}  & BLIP Diffusion~\cite{li2023blip} & 0.2934 &  0.7899&0.8620&0.5855&0.6692
\\
& ELITE~\cite{wei2023elite} & 0.2961 & 0.7924&0.8615&0.5922&0.6805
\\
& \jedi (1 input) & \textbf{0.3040} &  0.7818&0.8764&0.6190&0.7510
\\
\cmidrule{2-7}
& \jedi (3 inputs) & 0.2932 &  \textbf{0.8139} & \textbf{0.9011} & \textbf{0.6791} & \textbf{0.8037}
\\
\bottomrule
\end{tabular}
}
\vspace{-15pt}
\end{center}
\end{table*}

\textbf{Evaluation metrics. }The two main evaluation criteria for personalized text-to-image generation include (1) alignment between generated images and the input text prompts, and (2) faithfulness of the generations to the input reference images. 
We use the CLIP image-text similarity (CLIP-T) between the generated images and the input captions for (1). For (2), we follow prior works~\cite{ruiz2023dreambooth,li2023blip,wei2023elite} and use the cosine similarity of CLIP~\cite{radford2021learning} image embedding (CLIP-I) and DINO~\cite{caron2021emerging} image embedding (DINO) between the generated images and the reference images. DINO is considered to be a preferred metric for measuring image similarity as it is sensitive to the appearance variations of different images in the same concept class. Additionally, we also report CLIP-I and DINO scores only on the foreground masked images, i.e., images with foreground objects cutout using object detection and segmentation~\cite{liu2023grounding,kirillov2023segany}. This helps remove the background variations when computing the image similarity scores to better reflect the faithfulness to the reference subject. We call these metrics MCLIP-I and MDINO. 


\begin{table}[tb]
\renewcommand*{\arraystretch}{1}
\caption{\textbf{Effect of varying the S$^3$ dataset size. }}
\vspace{-15pt}
\label{tab:ablation_data}
\begin{center}
\setlength{\tabcolsep}{5pt}
\resizebox{0.95\linewidth}{!}{%
\begin{tabular}{l|p{15mm} p{15mm} p{15mm} p{15mm}}
\toprule
 & \multicolumn{4}{|c}{Dataset size} \\
 & \centering 0 & 0.6M & 1.2M & 1.6M\\
\cmidrule{2-5}
DINO ($\uparrow$) & 0.8895 &0.7501 &0.7443 &0.7467\\
MDINO ($\uparrow$) & 0.8931 & 0.8639 &0.8572 &0.8636\\
CLIP-T ($\uparrow$) & 0.2524 & 0.3020 &0.3017&0.3015\\
\bottomrule
\end{tabular}
}
\end{center}
\end{table}
\begin{table}[tb]
\renewcommand*{\arraystretch}{1}
\vspace{-5pt}
\caption{\textbf{Ablation of different joint-image diffusion designs. }}
\vspace{-15pt}
\label{tab:ablation_model}
\begin{center}
\setlength{\tabcolsep}{5pt}
\resizebox{\linewidth}{!}{%
\begin{tabular}{l|c|cccc}
\toprule
 & CLIP baseline & \jedi &(+) CLIP emb & w/o IG &  \\
\cmidrule{2-5}
DINO ($\uparrow$) & 0.3411 &0.7501&0.7432 & 0.4652\\
MDINO ($\uparrow$) & 0.4394 & 0.8639&0.8617 & 0.5922\\
CLIP-T ($\uparrow$) & 0.3325 & 0.3020&0.3041 & 0.3259\\
\bottomrule
\end{tabular}
}
\vspace{-15pt}
\end{center}
\end{table}

\textbf{Implementation details. }We implement our method based on StableDiffusion V1.4 to enable fair comparison with prior approaches. We train the model 
with batch size $2048$ and a learning rate $5e-5$. We initialize the weights using the pre-trained StableDiffusion model. For each batch, we randomly sample image sets from S$^3$, WebVid10M, Objaverse and LAION datasets with equal probability. We randomly choose the image set size between $2$-$4$ except for LAION, where the set size is always $1$. Our model takes $36$ hours to train on $32$ A100 GPUs for $140K$ steps. 

\subsection{Comparison}
We compare our method with the state-of-the-art finetuning-free methods - BLIPD~\cite{li2023blip} and ELITE~\cite{wei2023elite}, along with two finetuning-based methods - DreamBooth (DB)~\cite{ruiz2023dreambooth} and CustomDiffusion (CD)~\cite{kumari2022customdiffusion}. For each subject, we use one input reference image for finetuning-free approaches and three for finetuning-based approaches. 

\textbf{Comparison to finetuning-free methods. } Fig.~\ref{fig_cmp} shows the visual comparison of our results to BLIPD and ELITE. It can be seen that our method can faithfully capture the visual features of the input reference image, including both semantic attributes and low-level details. However, the generations from BLIPD and ELITE can only roughly resemble the color patterns and semantic features of the input. We also observe that BLIPD and ELITE can produce reasonable results for common object classes such as dogs. This is because their encoder can easily recognize the common object categories (such as the dog breed) which makes the personalized generation easier. However, for unique uncommon objects, their results tend to be much worse,~\eg the toy in the second row and the image in the third row. Note that even for the common classes such as the dog example in the first row, the generations from BLIPD and ELITE can miss some input features (different haircuts despite being the same breed). 
In contrast, our method eliminates information loss caused by the encoder and results in much better preservation of the custom concept. This is also reflected in the quantitative comparison Table~\ref{tab:comp}, where our method outperforms BLIPD and ELITE by a large margin. 

\textbf{Comparison to finetuning-based methods. } Fig.~\ref{fig_cmp2} shows the visual comparison of our approach with DreamBooth (DB) and CustomDiffusion (CD). When the number of reference images is limited, it is challenging for finetuning-based methods to avoid overfitting and generate novel variations of the input subject. From Fig.~\ref{fig_cmp2}, we see that DB often directly copies the input image due to overfitting, while the images generated by CD do not faithfully preserve the features of the input subject. In contrast, our method creates proper variations of the reference subjects without changing its key visual features. Even without any expensive finetuning, our method outperforms DB and CD in quantitative comparisons when we provide the same number of reference images (\jedi-3), as shown in Table~\ref{tab:comp}. 

\subsection{Ablation Study}

\textbf{Size of S$^3$ dataset. }Table~\ref{tab:ablation_data} reports the results of training our model using different numbers of synthetic images. Dataset size $0$ refers to training the model on only videos and multi-view images. This setting yields the best image alignment (DINO and MDINO) and the worst text alignment as the model learns a shortcut to ignore the text prompts and copy the input images. 
The performance of columns $2$-$4$ are roughly similar, which shows that we do not obtain much gains by increasing the size of the synthetic dataset. 


\textbf{Joint-image diffusion model. } In Table.~\ref{tab:ablation_model}, we report the ablation study of different design choices in the training of \jedi. The first column shows the results of a CLIP encoder baseline, which is a single-image diffusion model conditioned on the CLIP image features of the reference image. Our \jedi model yields a much better image alignment than the CLIP encoder baseline, which demonstrates the advantages of using the joint-image model over the image encoders. We also find that adding CLIP image embedding as extra conditional input to \jedi (+ CLIP emb) does not improve the performance as shown in column 3. This implies that the joint-image model already captures the information extracted in the image embedding. The last column reports the results without image guidance (w/o IG). By comparing the second and the last columns, we can see that image guidance is crucial to obtaining good personalization results.


\section{Conclusion}
This paper presents \jedi, a novel approach for finetuning-free personalized text-to-image generation using a joint-image diffusion model. We show how a single image diffusion U-Net can be adapted to learn a joint image distribution using coupled self-attention layers. To train the joint-image diffusion model, we construct a synthetic dataset called S$^3$, in which each sample contains a set of images sharing the same subject. The experimental results show that the proposed \jedi model outperforms the previous approaches both quantitatively and qualitatively in benchmark datasets. 
\setcounter{page}{1}
\maketitlesupplementary

\section{Implementation Details}
\label{sec:impl_supp}
In this section, we describe the key modifications based on StableDiffusion v1.4\footnote{We compare with previous approaches using the model based on v1.4 while the images in Fig.~1 are generated by the model based on SDXL.} to implement the proposed method. We point to the original location in StableDiffusion code and highlight the modified lines in each code snippet. 

\begin{figure}[h]
    \begin{center}
    \includegraphics[width=.9\linewidth]{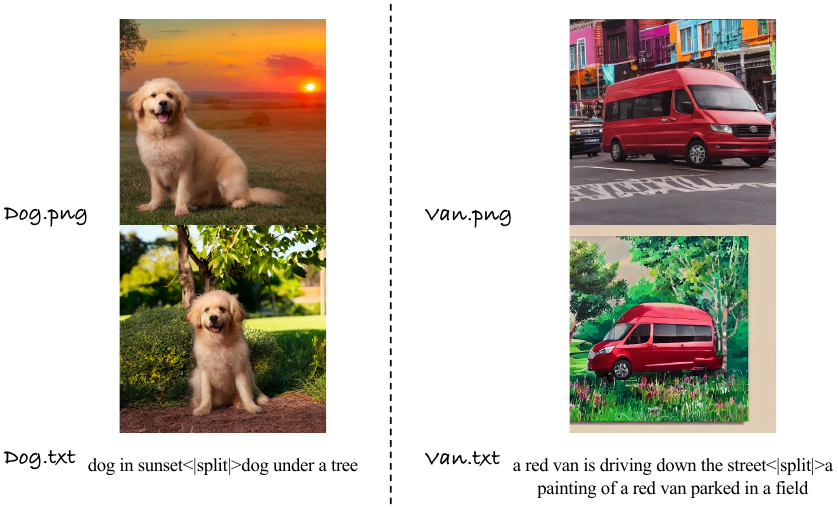}
    \end{center}
    \vspace{-15pt}
    \caption{\small \textbf{Data format.}
    An image file is the concatenation of all images in the set. The text file contains corresponding captions separated by a special token. }
    \vspace{-10pt}
    \label{fig_data_format}
\end{figure}

\textbf{Data loading. }We store all images and captions that belong to an image set in a single image file and text file. Images are vertically concatenated and the captions corresponding to different images are separated by a special token \texttt{<|split|>}, as illustrated in Fig.~\ref{fig_data_format}. This enables us to easily reuse existing single-image dataloaders in PyTorch. 
We only use square images in training and obtain the image set size by dividing the image height with the image weight. Given a batch of data, we extract individual images and text, and obtain the size of the image set \texttt{ng} as follows,
\begin{figure}[h!]
    \begin{center}
    \vspace{-10pt}
    \includegraphics[width=\linewidth]{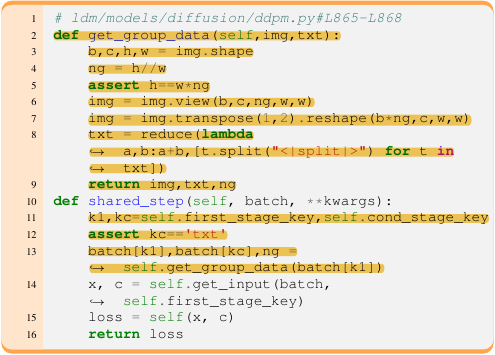}
    \end{center}
    \vspace{-10pt}
\end{figure}


\textbf{Joint-image diffusion models. }The proposed joint-image diffusion model can be easily implemented based on a single-image diffusion model with a few simple modifications as follows,
\begin{figure}[h!]
    \begin{center}
    \vspace{-10pt}
    \includegraphics[width=\linewidth]{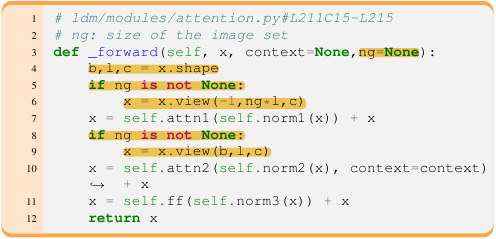}
    \end{center}
    \vspace{-10pt}
\end{figure}




\textbf{Personalization as inpainting. }As described in Sec.~3.3 of the paper, we cast the personalized generation problem into an inpainting task. In training, the inpainting masks are generated randomly. First, a binary random vector \texttt{ng\_mask} is sampled with every bit set to zero or one with equal probability. Then a \texttt{mask} of the same spatial size as the input image is constructed by replicating \texttt{ng\_mask} alone the height and width dimensions. The actual input fed into the U-Net is the concatenation of the mask \texttt{mask}, noisy image \texttt{x}, and masked clean image \texttt{x0*(1-mask)}. 
The key implementation can be found in the following code snippet. 
\begin{figure}[h!]
    \begin{center}
    \vspace{-10pt}
    \includegraphics[width=\linewidth]{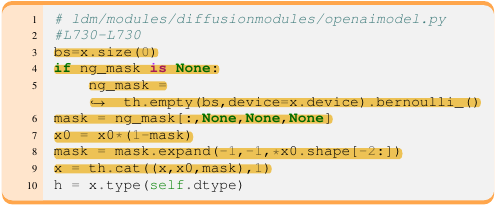}
    \end{center}
    \vspace{-10pt}
\end{figure}



\noindent At test time, \texttt{x0} is the concatenation of all input images and an all-zero image. The corresponding elements in \texttt{ng\_mask} are set to $0$ for the input images and $1$ otherwise. 

\begin{table*}[t]
\renewcommand*{\arraystretch}{1}
\vspace{-20pt}
\caption{Quantitative comparisons on the unique subject test set.}
\vspace{-15pt}
\label{tab:comp_supp}
\begin{center}
\setlength{\tabcolsep}{5pt}
\resizebox{.6\linewidth}{!}{%
\begin{tabular}{l|cccccc}
\toprule
 Method &CLIP-T ($\uparrow$) & CLIP-I ($\uparrow$) &MCLIP-I ($\uparrow$) & DINO ($\uparrow$) & MDINO ($\uparrow$)\\
\midrule
BLIP Diffusion~\cite{li2023blip} & 0.2851 &  0.8107&0.8234&0.6091&0.6018
\\
ELITE~\cite{wei2023elite} &0.2193  & 0.6082&0.6430&0.1862&0.2156
\\
\jedi & \textbf{0.2856} &  \textbf{0.8697}&\textbf{0.8838}&\textbf{0.7934}&\textbf{0.7926}
\\
\bottomrule
\end{tabular}
}
\vspace{-15pt}
\end{center}
\end{table*}
\textbf{Synthetic same-subject (S$^3$) dataset. }Algorithm~\ref{alg:1} describes the training data generation (Fig.~3 in the paper) in details. Please note that we use the term \textit{instance segmentation} for simplicity; however, in our implementation, we combine an object detection model~\cite{liu2023grounding} and a segmentation model~\cite{kirillov2023segment} to separate object instances rather than using an actual instance segmentation model. The \textit{object-centric} prompts (GPT($l,object$) in Algorithm~\ref{alg:1}) are generated by instructing ChatGPT to generate details of an object $l$, and the \textit{scene-centric} prompts (GPT($l,scene$)) in Algorithm~\ref{alg:1}) are generated by instructing it to describe a scene involving the object $l$. The list of object names used in our implementation and a random subset of the generated training samples can be found in the attached file. We found that the initial text prompts generated by ChatGPT lack variations and therefore pair the images with the captions obtained from BLIPv2~\cite{li2023blip} in training samples.
\begin{algorithm}[h]
\caption{Generating the S$^3$ dataset.  }\label{alg:1}
\begin{algorithmic}[1]
\STATE {\bfseries Input:} A list of object names $L$;
\STATE ~~~~~~~~~~~ A text-to-image model $G$; 
\STATE ~~~~~~~~~~~ A text-based inpainting model $G_I$; 
\STATE ~~~~~~~~~~~ ChatGPT GPT; 
\STATE ~~~~~~~~~~~ An instance segmentation model $S$; 
\STATE ~~~~~~~~~~~ CLIP image model CLIP; 
\STATE {\bfseries Output:} a database $\mathcal{X}$ of image sets $\mathcal{X}=\{X_1,X_2,...,X_P\}$ where $X_p=\{x_1,x_2,...,x_{N_p}\}$ is a set of images share a common subject;
\STATE $\mathcal{X}\leftarrow \phi$;
\FOR {$l$ {\bfseries in} $L$}
\STATE Generate an object-centric prompt $t_o \leftarrow \mbox{GPT}(l,object)$;
\STATE Generate an image $x_0 \leftarrow G(t_o)$; 
\STATE Extract object instances from $x_0$: $\{\hat{x}_1,\hat{x}_2,...,\hat{x}_K\} \leftarrow S(x_0)$;
\STATE Construct an affinity matrix $A, A_{ij}\leftarrow\mbox{CLIP}(\hat{x}_i,\hat{x}_j)$;
\STATE Construct an adjacent matrix $M, M_{ij}\leftarrow 1$ if $A_{ij}>0.95$ else $0$;
\STATE Find connected components $\mathcal{I}\leftarrow \{I_1,I_2,...,I_P\}$ of the graph represented by $M$;
\FOR {$I_p$ {\bfseries in} $\mathcal{I}$ }
    \STATE $X \leftarrow \phi$;
    \FOR {$i$ {\bfseries in} $I_p$ }
    \STATE Generate a scene-centric prompt involving object name $l$: $t_s\leftarrow \mbox{GPT}(l,scene)$;
    \STATE Paste the object instance image $\hat{x}_i$ at a random location in an empty image $x$;
    \STATE Inpaint the unknown area in $x$: $x\leftarrow G_I(x,t_s)$;
    \STATE $X \leftarrow X\cup \{x\}$;
\ENDFOR
\STATE $\mathcal{X} \leftarrow \mathcal{X}\cup \{X\}$
\ENDFOR
\ENDFOR
\end{algorithmic}
\end{algorithm}
\vspace{-10pt}

\textbf{Evaluation metric. }We use the CLIP ViT-B/32 to compute CLIP-T, CLIP-I and MCLIP-I. We use DINO VIT-S/16 to compute DINO and MDINO. We use one input image per subject for comparison with finetuning-free methods, and average the pair-wise scores to all real images of the same subject and over all possible choices of the input image. For comparison with finetuning-based methods, we randomly select three input images. In ablation studies we use one randomly selected input image for each subject and only compute the scores using input/output pairs by default unless stated otherwise. 

\section{Additional Experiments}
\subsection{Additional Results}
Fig.~\ref{fig:our_3in_suppl} shows additional personalized generation results on real-world human and object images. Our method can generate high-quality images with diverse content while preserving the key visual features of the subjects in input images. Although the model is not trained on human-specific data, it can still generate reasonable results for human subjects, as shown in the second and third row of Fig.~\ref{fig:our_3in_suppl}. 

\subsection{Comparison with State-of-the-Art Methods}
Fig.~\ref{fig_cmp2_suppl} provides additional visual comparisons with finetuning-based methods DreamBooth (DB) and CustomDiffusion (CD). Finetuning-based methods suffer from the overfitting issue and might fail to preserve the subject identity.  For common subjects, they tend to extensively copy from the reference images, adding only minor adjustments to match the given text,~\eg in the first example, for the prompt \textit{a backpack in the snow}, DreamBooth nearly replicate a reference image with slight snow patterns added in the bottom. For unique subjects, the finetuning-based methods often fail to preserve the distinctive features,~\eg the cartoon character in the fourth row.  This is because these methods use the loss on retrieved or generated images of similar subjects as regularization during finetuning. For unique and rare objects, these images can be visually distinct from the reference images and interfere the model from memorizing the custom concept. 

To further demonstrate the advantage of our methods for challenging cases, we collect a new test set containing unique subjects with only single input image for each subject. Most the input images are from Reddit AI Art channel\footnote{https://www.reddit.com/r/aiArt/}. Fig.~\ref{fig_testinput} visualize the input images.  The first five rows in Fig.~\ref{fig_cmp_suppl} compare the results of our method to state-of-the-art finetuning-free methods BLIP-Diffusion (BLIPD) and ELITE on the unique subject test set. We also include the results on common subjects from DreamBooth test set in the last two rows for comparison. It can be seen that BLIPD and ELITE can produce reasonable results for common subjects such as the dog of a typical breed and a common stuffed animal (row 6-7). However, for unique subjects, their results hardly resemble the subject from the reference image (row 1-5). In contrast, our method can faithfully capture the key visual features of the subject. The advantage of our method is also clearly reflected in the quantitative results in Table~\ref{tab:comp_supp}, where our method outperforms ELITE and BLIP-Diffusion by a large margin. 

\subsection{Additional Analysis}
\textbf{Image guidance. }As we have discussed in the paper, the use of image guidance can significantly improve the faithfulness to the input images. This is also supported by the visual comparison in Fig.~\ref{fig:ablation_suppl} (column 4-5). 

In our main experiments in the paper, we use a simple strategy for image guidance where both the image and text input are set to null for unconditional inference. Here we discuss a more flexible guidance strategy to model trade-off between image alignment and text alignment. The score function with flexible image guidance is as follows,
\begin{equation}
\tilde{\bm{\epsilon}}(\bm{x}_t,\bm{\hat{x}},\bm{M})=\bm{\epsilon}^0 + \lambda_1(\bm{\epsilon}^1 -\bm{\epsilon}^0) + \lambda_2[\bm{\epsilon}_\theta(\bm{x}_t,\hat{\bm{x}},\bm{M})-\bm{\epsilon}^1],
\end{equation}
where $\bm{\epsilon}^0=\bm{\epsilon_\theta}(\bm{x}_t,\bm{0},\bm{M})$ represents the unconditional score when the text prompt and all reference images are set to null; $\bm{\epsilon}^1$ represents the partially conditional score when either the text prompt or reference images are kept. We can compute the partial conditional score $\bm{\epsilon}^1$ using image conditioning to emphasize text alignment, or text conditioning to emphasize image alignment. We call these two options \emph{text first} and \emph{image first} strategies, respectively. Table~\ref{tab:guide} reports the quantitative results based on different strategies averaged over a varying guidance scale in $[1.5,10]$. It can be seen that the \emph{text first} strategy yields higher DINO and MDINO scores, indicating better image alignment. \emph{Image first} strategy yields a higher CLIP-T score, which indicates better text alignment. 
 \begin{table}[h]
\caption{Quantitative results with different guidance strategies averaged over a varying guidance scale in $[1.5,10]$.
}
\vspace{-10pt}
\label{tab:guide}
\begin{center}
\resizebox{.8\columnwidth}{!}{
\setlength{\tabcolsep}{5pt}
\begin{tabular}{c|cccc}
\toprule
Strategy & Text only & Joint & Image first& Text first\\
\midrule
DINO & 0.4652 &0.7268 &0.6558 &0.7508\\
MDINO & 0.5922 & 0.8384 &0.7863&0.8527\\
CLIP-T & 0.3259 & 0.3013 &0.3156&0.2853\\
\bottomrule
\end{tabular}}
\vspace{-10pt}
\end{center}
\end{table}
\begin{figure}[t!]
\begin{center}
\vspace{-20pt}
\includegraphics[width=.9\linewidth]{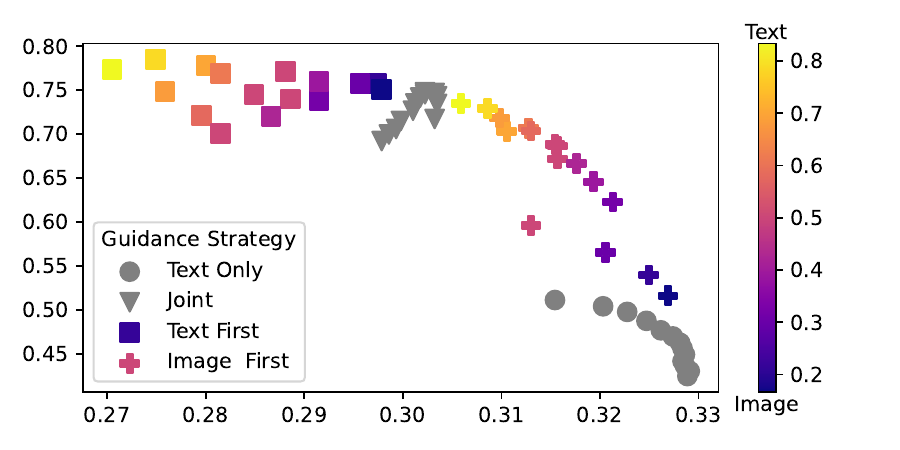}
\end{center}
\vspace{-20pt}
\caption{\small Effect of varying the ratio between the guidance scales for image and text guidance. X-axis: CLIP-T; Y-axis: DINO. }
\vspace{-10pt}
\label{fig_ablation_guide}
\end{figure}

We can also adjust the ratio between the guidance scale of image condition and text condition for more flexible personalized generation. Fig.~\ref{fig_ablation_guide} visualize the change of DINO and CLIP-T scores with the varying ratio. 
We can see that the use of image guidance is important. Using only text guidance (Text only in Fig.~\ref{fig_ablation_guide}) yields low DINO score, indicating low resemblance to the custom subject. By varying the ratio of text and image guidance scales we can balance between subject identity preservation and text alignment. We found that the simple joint guidance strategy (Joint in Fig.~\ref{fig_ablation_guide}) usually gives good balanced results. 

 \begin{table}[h]
\caption{Quantitative comparison with three baseline models.  
}
\vspace{-10pt}
\label{tab:ablation_suppl}
\begin{center}
\resizebox{\columnwidth}{!}{
\setlength{\tabcolsep}{5pt}
\begin{tabular}{c|cccc}
\toprule
Model& CLIP baseline & Concate. baseline & Learned baseline & \jedi \\
\midrule
DINO & 0.3411 &0.3379 & 0.7065 &0.7501 \\
MDINO & 0.4394 & 0.4292 &0.7740 &0.8639\\
CLIP-T & 0.3325 & 0.3247 &0.3015 &0.3020\\
\bottomrule
\end{tabular}}
\vspace{-10pt}
\end{center}
\end{table}

\textbf{Comparison with baseline models. }We have discussed a CLIP-encoder baseline in the paper (CLIP baseline). Here we include the comparison to another two baseline models. Concate baseline: the input images are concatenated to the noisy images to be fed into the UNet. Learned baseline: similar to SuTI~\cite{chen2024subject} where an learnable encoder is used to extract a feature vector from the input images. All models are trained on the same training data described in Sec.~4 of the paper and are based on the same StableDiffusion v1.4 backbone for fair comparison. Table~\ref{tab:ablation_suppl} reports the quantitative comparison results and the visual comparison can be found in Fig.~\ref{fig:ablation_suppl} (column 1,2,3,5). The results indicate that our method significantly outperforms all baseline models in terms of image alignment, as evidenced by considerably higher DINO and MDINO scores. 

\textbf{Training data identity similarity. } 
Table~\ref{tab:data_sim} reports the average CLIP and DINO scores on training samples over the 1,000 ImageNet categories, which indicates a high overall identity similarity for a wide array of categories. For context, we also show the scores on real images from the DreamBooth test set, which has a slightly higher identity similarity but covers much less categories than our dataset. 
\begin{table}[h]
\renewcommand*{\arraystretch}{1}
\begin{center}
\small
\begin{tabular}{ccccc}
\toprule
& Subjects & Categories & CLIP-I ($\uparrow$) & DINO ($\uparrow$)\\
\midrule
S$^3$ dataset & $1.6M$ & $\sim2K$ & 0.849 &0.751 \\
Real images & 30 & 15 & 0.885 & 0.774 \\
\bottomrule
\end{tabular}
\end{center}
\caption{Training data statistics. }
\label{tab:data_sim}
\end{table}

\noindent\textbf{Quantitative results with more input images. }Although our method does not have an inherent constraint on the number of inputs, for simplicity, we only use 1-3 input images in the current implementation. We find that our method still outperforms DB and CD, even when they are finetuned with the maximum available reference images in the test set (4-6 images), as shown in Table~\ref{tab:r3}. We will add the experiments with more reference images, \eg 10, in the revised version. 
\begin{table}[h]
\caption{\scriptsize Comparison to DB and CD with the maximum available reference images. }
\label{tab:r3}
\begin{center}
\setlength{\tabcolsep}{5pt}
\resizebox{\linewidth}{!}{
\begin{tabular}{c|ccccc}
\toprule
 &CLIP-T & CLIP-I ($\uparrow$) &MCLIP-I ($\uparrow$) & DINO ($\uparrow$) & MDINO ($\uparrow$)\\
\midrule
DB&0.2971&0.8025 & 0.8736 & 0.6226 & 0.7175\\
CD&0.3071&0.7864 & 0.8586 & 0.6198 & 0.7011\\
\midrule
Ours (1 input) &0.3040&0.7818 & 0.8764 & 0.6190 & 0.7510\\
Ours (3 inputs) &0.2932&0.8139 & 0.9011 & 0.6791 & 0.8037\\
\bottomrule
\end{tabular}
}
\end{center}
\end{table}

\noindent\textbf{Inference cost. }The inference cost is comparable to other methods when $N$ is small (reported in the table below). When $N$ is substantially larger,~\eg a database, we can reduce the inference cost by first finetuning the model on the database, and then retrieving the few images closest to the text prompt to be the actual test time input (please refer to the future work section). 
\begin{table}[h]
\caption{\scriptsize Inference time for one diffusion step on one A100 GPU. }
\label{tab:r4}
\begin{center}
\setlength{\tabcolsep}{5pt}
\resizebox{.6\linewidth}{!}{
\begin{tabular}{c|ccc}
\toprule
Method & BLIPD & ELITE & Ours\\
\midrule
Time (second) $\downarrow$ & 0.0492 & 0.0719 & 0.0564\\
\bottomrule
\end{tabular}
}
\end{center}
\end{table}

\section{Limitations and Future Work}
A limitation of \jedi is that it needs to process all reference images at inference time. This enables finetuning-free personalization but leads to efficiency drop when the number of reference images increases. Therefore, \jedi is more suitable for subject image generation given a few reference images, and are less efficient in adapting to a new domain given a large database of reference images. A potential solution is to combine \jedi with finetuning-based methods. When a large database of reference images are available, we can first finetune \jedi on the database. Then at inference time, given a text prompt, we retrieve the most relavent images from the database to use as the test-time inputs to \jedi. Another limitation is that the current implementation cannot be directly applied for multi-subject image generation. There are two possible ways to extend \jedi for multi-subject generation: (1) generate multiple subjects sequentially through inpainting, and (2) construct a multi-subject S$^3$ dataset by combining multiple sets of subjects. We will explore these directions in future work. 
\setlength{\tabcolsep}{0.5pt}
\renewcommand{\arraystretch}{0.5}
\begin{figure*}[ht!]
    \centering
    \vspace{-20pt}
    \includegraphics[width=\textwidth]{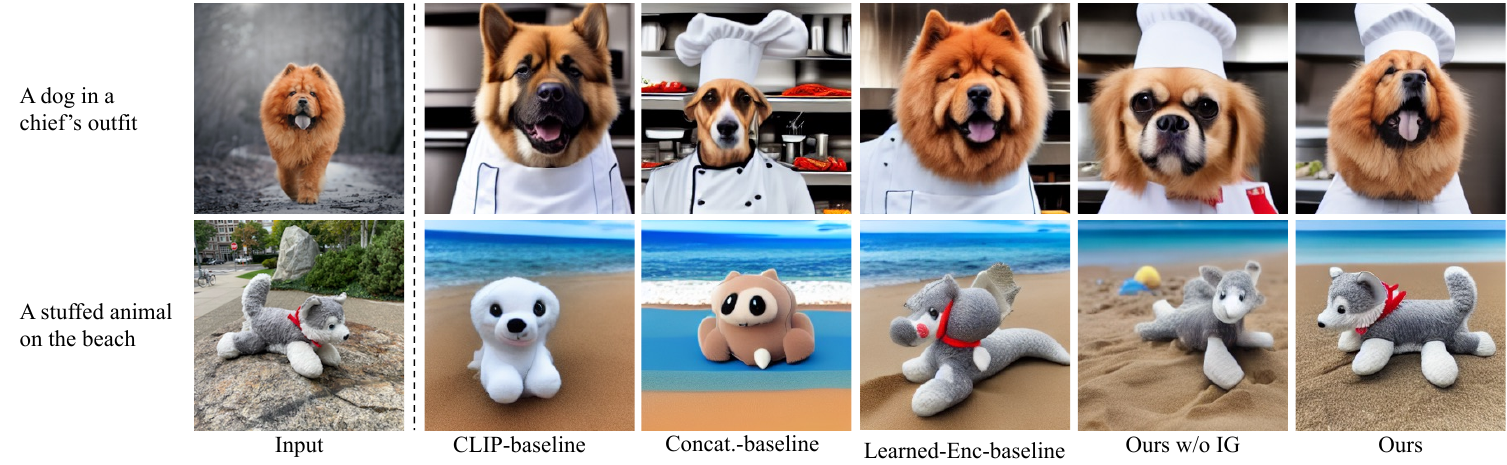}
    \vspace{-10pt}
    \caption{Visual comparisons to baseline models.  }
    \label{fig:ablation_suppl}
\end{figure*}
\setlength{\tabcolsep}{0.5pt}
\renewcommand{\arraystretch}{0.5}
\begin{figure*}[ht!]
    \centering
    \vspace{-10pt}
    \includegraphics[width=\textwidth]{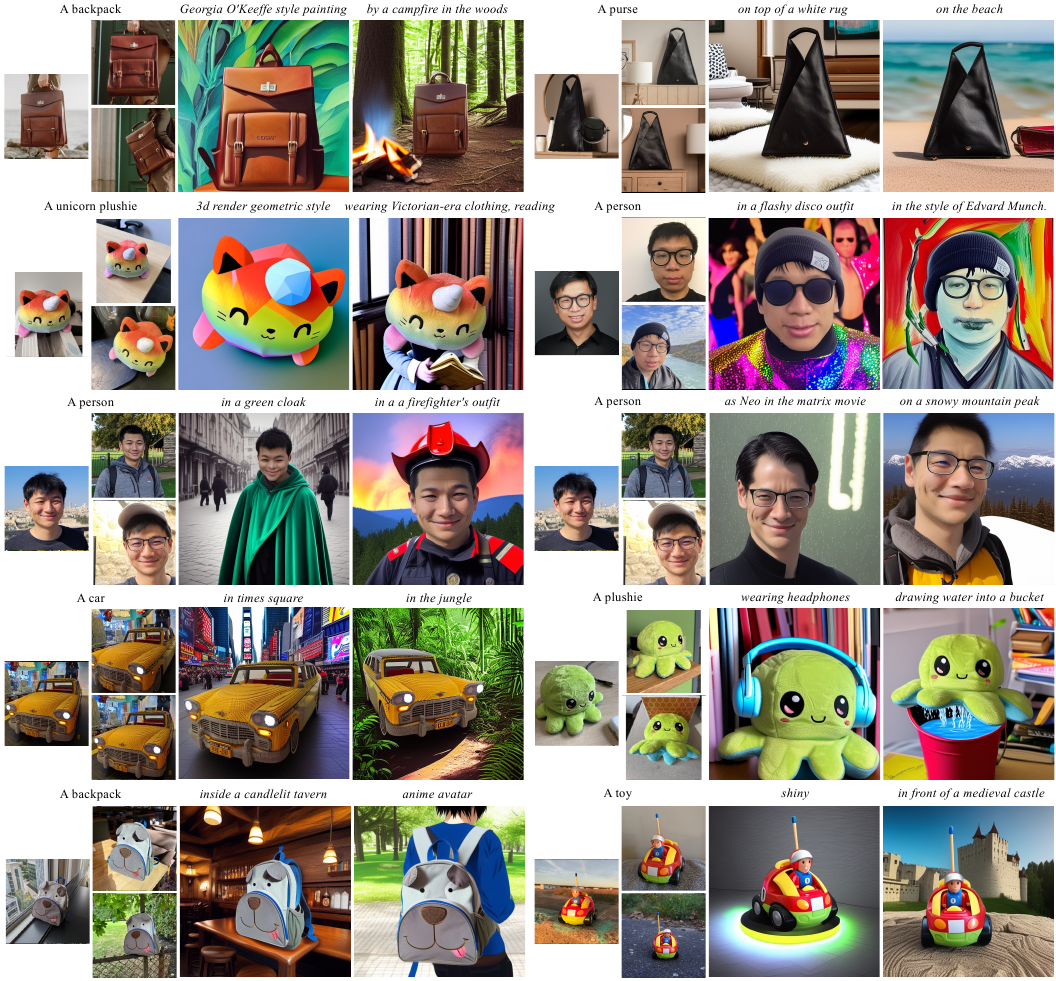}
    \vspace{-5pt}
    \caption{Personalized generation results for human and real-world objects. }
    \vspace{-20pt}
    \label{fig:our_3in_suppl}
\end{figure*}

\begin{figure*}[t]
    \begin{center}
    \includegraphics[width=\linewidth]{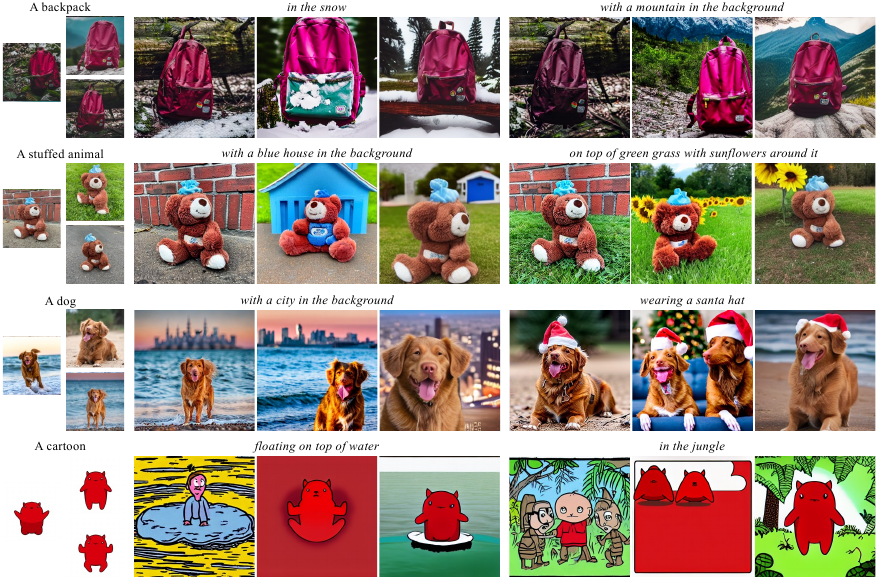}\\
    \scriptsize{ \hfill {Input} \hfill\hfill {DB} \hfill\hfill {CD} \hfill\hfill {Ours} \hfill\hfill {DB} \hfill\hfill {CD} \hfill\hfill {Ours} \hfill\hfill}
    \end{center}
    \vspace{-15pt}
    \caption{Visual comparison with finetuning-based methods on DreamBooth test set. DreamBooth (DB) and CustomDiffusion (CD) tend to overfit for common subjects and fail to capture the visual features of unique subjects. }
    \label{fig_cmp2_suppl}
\end{figure*}
\begin{figure*}[t]
    \begin{center}
    \vspace{-0pt}
    \includegraphics[width=\linewidth]{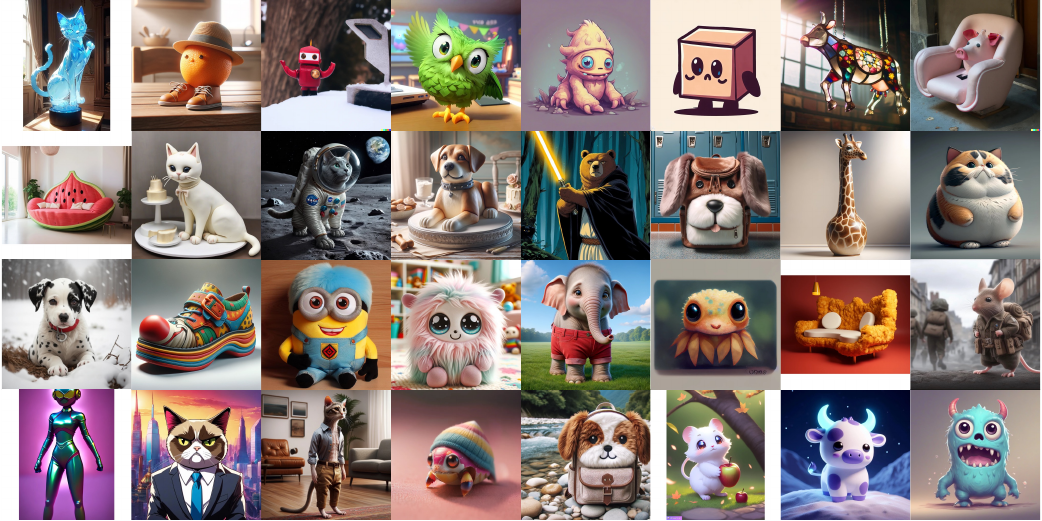}
    \end{center}
    \vspace{-15pt}
    \caption{Input images in the unque subject test set.}
    \vspace{-20pt}
    \label{fig_testinput}
\end{figure*}
\begin{figure*}[t]
    \begin{center}
    \vspace{-10pt}
    \includegraphics[width=\linewidth]{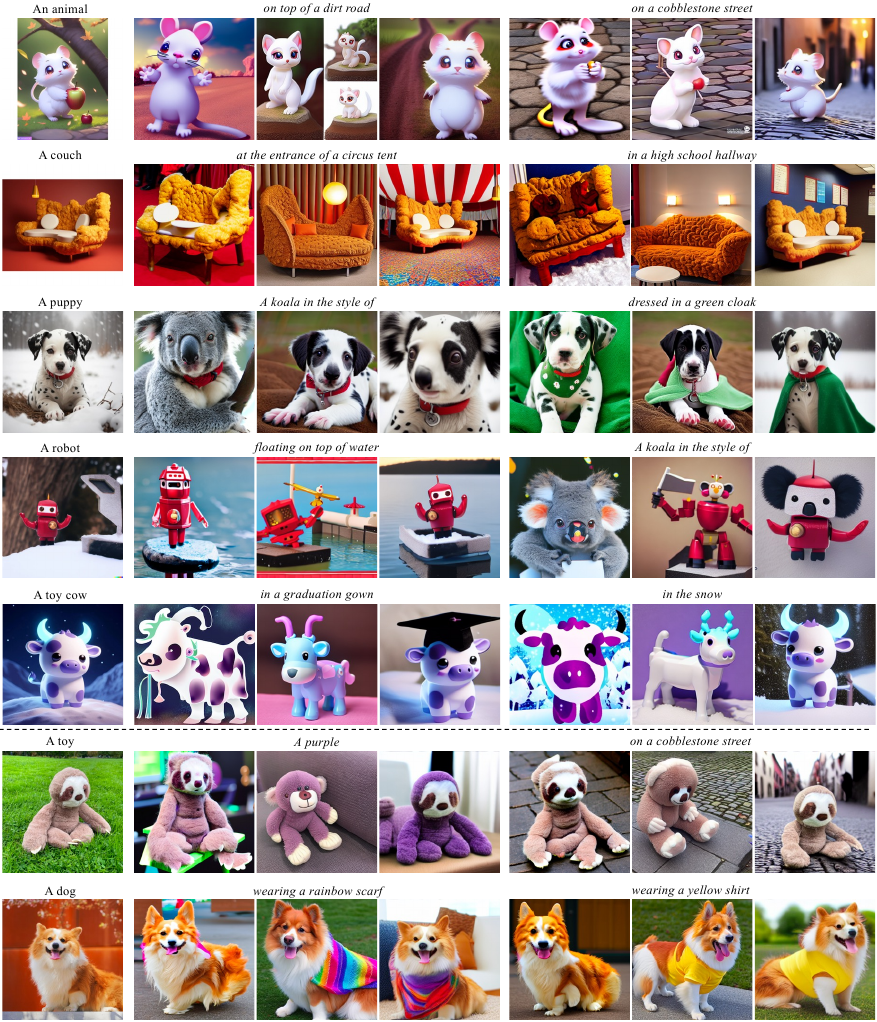}\\
    \scriptsize{ \hfill {Input} \hfill\hfill {ELITE} \hfill\hfill {BLIPD} \hfill\hfill {Ours\textcolor{white}{1}} \hfill\hfill {ELITE} \hfill\hfill {BLIPD} \hfill\hfill {Ours\textcolor{white}{1}} \hfill\hfill}
    \end{center}
    \vspace{-15pt}
    \caption{Visual comparison with finetuning-free methods on the unique subject test set (row 1-5) and on Dreambooth test set (row 6-7). BLIP Diffusion and ELITE can generate reasonable results for common subjects but often fail in challenging cases involving unique subjects. In contrast, our method can handle challenging cases and generate personalized images with well-preserved details. }
    \vspace{-0pt}
    \label{fig_cmp_suppl}
\end{figure*}

{
    \small
    \bibliographystyle{ieeenat_fullname}
    \bibliography{main}
}


\end{document}